%% file: OC-MAINS-ext.tex
\documentclass[conference]{IEEEtran}

\usepackage{amsthm}
\usepackage{cite}
\usepackage{amsmath,amssymb,amsfonts}
\usepackage{algorithmic}
\usepackage{graphicx}
\usepackage{textcomp}
\usepackage{wrapfig}
\usepackage{mleftright}
\usepackage{xparse}
\usepackage{ifthen}

\newcommand{\extendedversion}{1}

\NewDocumentCommand{\evalat}{sO{\big}mm}{%
  \IfBooleanTF{#1}
   {\mleft. #3 \mright|_{#4}}
   {#3#2|_{#4}}%
}

\DeclareMathOperator*{\argmin}{arg\,min}
\usepackage{color, colortbl}
\definecolor{lightgray}{rgb}{0.83, 0.83, 0.83}
\usepackage{tikz}
\usepackage[hidelinks]{hyperref}
\usepackage{algorithm,algorithmic}
\usepackage[flushleft]{threeparttable}
\usepackage{hyperref}
\usepackage{textcomp}
\usepackage{subcaption}

\usepackage[inline]{enumitem}
\usepackage{mathtools,leftindex,tensor,mhchem}
\makeatletter
\newcommand{\subalign}[1]{%
  \vcenter{%
    \Let@ \restore@math@cr \default@tag
    \baselineskip\fontdimen10 \scriptfont\tw@
    \advance\baselineskip\fontdimen12 \scriptfont\tw@
    \lineskip\thr@@\fontdimen8 \scriptfont\thr@@
    \lineskiplimit\lineskip
    \ialign{\hfil$\m@th\scriptstyle##$&$\m@th\scriptstyle{}##$\hfil\crcr
      #1\crcr
    }%
  }%
}
\makeatother

\usepackage{caption}
\usepackage{subcaption}
\begin{document}

\ifthenelse{\equal{\extendedversion}{1}}{
  \title{An Observability-Constrained Magnetic Field-Aided Inertial Navigation System --- Extended Version}
}
{
\title{An Observability-Constrained Magnetic Field-Aided Inertial Navigation System}
}

\author{\IEEEauthorblockN{Chuan Huang}
\IEEEauthorblockA{\textit{Dept. of Electrical Engineering} \\
\textit{Linköping University}\\
Linköping, Sweden \\
chuan.huang@liu.se}
\and
\IEEEauthorblockN{Gustaf Hendeby}
\IEEEauthorblockA{\textit{Dept. of Electrical Engineering} \\
\textit{Linköping University}\\
Linköping, Sweden \\
gustaf.hendeby@liu.se}
\and
\IEEEauthorblockN{Isaac Skog}
\IEEEauthorblockA{\textit{Div. of Communication Systems} \\
\textit{KTH Royal Institute of Technology}\\
Stockholm, Sweden\\
skog@kth.se}
}

\maketitle

\begin{abstract}
Maintaining consistent uncertainty estimates in localization systems is crucial as the perceived uncertainty commonly affects high-level system components, such as control or decision processes. A method for constructing an observability-constrained magnetic field-aided inertial navigation system is proposed to address the issue of erroneous yaw observability, which leads to inconsistent estimates of yaw uncertainty. The proposed method builds upon the previously proposed observability-constrained extended Kalman filter and extends it to work with a magnetic field-based odometry-aided inertial navigation system. The proposed method is evaluated using simulation and real-world data, showing that (i) the system observability properties are preserved, (ii) the estimation accuracy increases, and (iii) the perceived uncertainty calculated by the EKF is more consistent with the true uncertainty of the filter estimates.
 \end{abstract}

\bstctlcite{IEEEexample:BSTcontrol}

\section{Introduction}
Maintaining consistent uncertainty estimates in localization systems is crucial as the perceived uncertainty commonly affects high-level system components, such as control or decision processes. In an odometry-aided inertial navigation system (INS), such as the magnetic field-odometry-aided INSs presented in~\cite{huang2023mains, zmitri2020magnetic,ZHANG2023MagODO,dorveaux2011combining}, the uncertainty about the position and yaw of the navigation platform can, if the errors in the initial state are uncorrelated, never become smaller than the initial position and yaw uncertainty. This is because odometry and inertial measurements only provide relative motion information. Hence, for any algorithm used to estimate the navigation state $x_k$ in an odometry-aided INS it should hold that
\begin{equation}\label{eq:inequalities}
    P^{\text{\footnotesize p}}_k\succeq P^{\text{\footnotesize p}}_0\quad \text{and}\quad  P^{\phi}_k\succeq P^{\phi}_0,
\end{equation}
where $P^{\text{\footnotesize p}}_k$ and $P^{\phi}_k$ denote the covariance of the, at time $k$, estimated position and yaw states, respectively. 

\begin{figure}[tb!]
\includegraphics[width=\columnwidth]{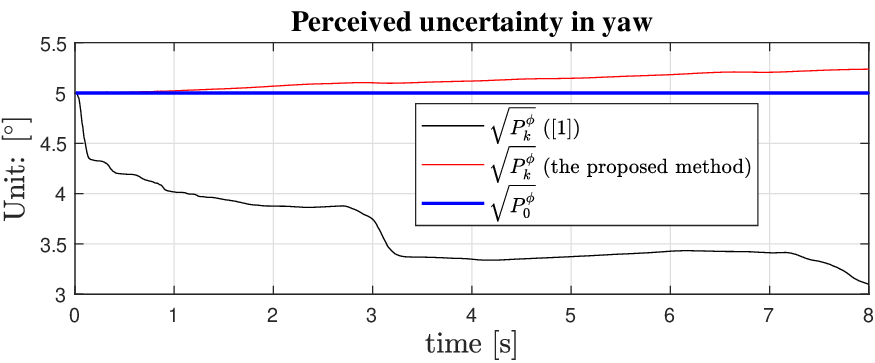}
\caption{Example of the perceived yaw uncertainty calculated by the EKF used to realize magnetic field-odometry-aided INS in \cite{huang2023mains} (black line). Also shown is the perceived yaw uncertainty of the proposed observability-constrained EKF algorithm (red line), as well as the initial uncertainty (blue line).}
\label{f: posterior vs prior uncertainty}
\end{figure}

In \cite{huang2023mains}, an extended Kalman filter (EKF) was used in a magnetic field-aided INS (MAINS). In Fig.~\ref{f: posterior vs prior uncertainty}, the square roots of $P^\phi_0$ and $P^\phi_k$ as calculated by the EKF in the MAINS are shown. As seen from the figure the square root of $P^\phi_k$ falls below that of $P^\phi_0$. Hence the inequality in \eqref{eq:inequalities} does not hold and the uncertainty estimate of the EKF is inconsistent. The inconsistency is because the EKF linearizes the system model around the estimated state, which causes the yaw to be perceived as observable even though it is not. This type of inconsistency effect has been observed in multiple EKF-based implementations of odometry-aided INS, see e.g., \cite{mingyang2013HighPrecision, Hesch2014Consistency, huang2011observability}, and is troublesome if the yaw information should be fused with information from other systems or be used in a control or decision process. Next, we will present a way to fix this inconsistency by modifying the EKF algorithm so that it preserves the observability properties of the underlying nonlinear system. The presented method is an extension of the method published in \cite{Hesch2014Consistency}. Our contributions are: 
\begin{itemize}
    \item An extension of how the basis vectors spanning the subspace of the unobservable state space (hereafter referred to as the unobservable subspace) can be chosen, simplifying the preservation of the observability properties.  
    \item The derivation of the unobservable subspace for the model used in \cite{huang2023mains}.
    \item An experimental evaluation that verifies the effectiveness of the proposed method when applied to the MAINS. 
\end{itemize}
The code and data to produce the presented results can be found at: \url{https://github.com/Huang-Chuan/OC-MAINS-code}.

\section{Preserve Observability Properties} \label{chp: preserve estimability}
The behavior of odometry-aided INS is commonly described by a nonlinear state-space model of the following form 
\begin{equation}
  \label{eq: odometry nonlinear state-space model}
  \begin{aligned}
    x_{k+1} &= f(x_k, u_k, w_k),  && \quad x \in \mathbb{R}^{n_x},\\
    y_k &= h(x_k) + e_k, &&\quad y \in \mathbb{R}^{n_y}.    
  \end{aligned}
\end{equation}
Here $f$ and $h$ are the nonlinear functions, and $u_k$ denotes the control input. Further, $w_k$ and $e_k$ denote the process and measurement noise, respectively. They are assumed to be white noise with covariance $Q_k$ and $R_k$, respectively.

Several approaches have been proposed to preserve observability properties of the model \eqref{eq: odometry nonlinear state-space model} in EKF filtering, see \cite{Hesch2014Consistency, HuaiZheng2018Robocentric, Brossard2019Exploiting, mingyang2013HighPrecision, huang2010observability, Caruso2019Magneto}, one of which is called the observability-constrained EKF \cite{Hesch2014Consistency}. The basic idea of the method in \cite{Hesch2014Consistency} is to modify the Jacobians used in the EKF, such that the basis of the unobservable subspace, evaluated at state estimates, lies in the nullspace of the observability matrix evaluated at state estimates of EKF. The unobservable subspace is a subspace of the nullspace of the local observability matrix \cite{Chen1990Local} associated with \eqref{eq: odometry nonlinear state-space model}, i.e. the observability matrix constructed as 
\begin{subequations}
\begin{equation} 
\label{eq: obs matrix at the nominal trajectory}
    \mathcal{\bar{O}}_{k} \triangleq 
    \begin{bmatrix}
      \bar{H}_{k} \\
      \bar{H}_{k + 1} \bar{\Phi}(k+1,k) \\
      \vdots \\
      \bar{H}_{k + n_x - 1} \bar{\Phi}(k+n_x-1,k)
    \end{bmatrix}.
\end{equation}
Here $\bar{x}_{k:k+n_x-1}$ denotes the nominal trajectory, which is the solution of  \eqref{eq: odometry nonlinear state-space model} with the control sequence ${\bar{u}_{k:k+n_x-1}}$ and the process noise turned off. Further,
\begin{equation}
    \bar{\Phi}(k +i,k)= \bar{F}_{k+i-1}\bar{F}_{k+i-2} \cdots \bar{F}_{k}
\end{equation}
where
\begin{equation}
    \bar{H}_{k} = \evalat[\bigg]{\frac{\partial h}{\partial x_{k}}}{x_{k}=\bar{x}_{k}}\quad\text{and}\quad
    \bar{F}_{k}=\evalat[\bigg]{\frac{\partial f}{\partial x_{k}}}{\subalign{x_{k}&=\bar{x}_{k}\\u_{k}&=\bar{u}_{k}\\w_k&=0}}.
\end{equation}
\end{subequations}
The unobservable subspace $\mathcal{N}_{k}$ can be represented as  
\begin{subequations}
\begin{equation}
\begin{split}
    \mathcal{N}_{k} &= \text{span}\{ N(\bar{x}_k)\}, 
\end{split}
\end{equation}
\end{subequations}
where $N: \mathbb{R}^n \rightarrow \mathbb{R}^{n \times p}$ is a matrix valued function such that $\bar{\mathcal{O}}_k N(\bar{x}_k)=0$, and $p$ denotes the dimension of the unobservable subspace. That is, the columns of $N(\bar{x}_k)$ are basis vectors that span the unobservable subspace. The basis of the unobservable subspace is not unique. Hence, right multiplying $N(\bar{x}_k)$ with any full-rank matrix $\mathcal{E}_k \in \mathbb{R}^{p\times p}$ will not change the span of the basis. That is, 
\begin{subequations}
\begin{equation}
    \text{span}\{ N^{\dagger}(\bar{x}_k)\}=\text{span}\{ N(\bar{x}_k)\} ,
\end{equation}
where 
\begin{equation}
    N^{\dagger}: \quad \mathbb{R}^n \rightarrow \mathbb{R}^{n \times p}\quad\text{and}\quad
    N^{\dagger}(x_k) = N(x_k) \mathcal{E}_k.
\end{equation}
\end{subequations}
This property, which was not used in \cite{Hesch2014Consistency}, will be important when modifying the Jacobians, as it reduces the changes made to the Jacobians to preserve the observability properties.

Once the basis vectors of the unobservable subspace are determined, they are, as will be described next, used to modify the Jacobians used in the EKF. Let 
\begin{equation}\label{eq:F and H normal}
        \hat{F}_k = \evalat[\bigg]{\frac{\partial f}{\partial x_k}}{\subalign{x_k&=\hat{x}_{k|k}\\u_k&=\hat{u}_k\\w_k&=0}} \quad\text{and}\quad
        \hat{H}_k = \evalat[\bigg]{\frac{\partial h}{\partial x_k}}{x_k=\hat{x}_{k|k-1}}
\end{equation}
be the Jacobians used in the unmodified EKF.  Here $\hat{x}_{k|k}$ and $\hat{x}_{k|k-1}$ denote the posterior and prior estimate of the state, i.e., the estimate of $x_k$ given measurements up to time $k$ and $k-1$, respectively. These Jacobians are modified by making the smallest possible (in terms of the Frobenius norm) changes to their entries while still preserving the observability properties. This is done by solving the following optimization problems.  
\begin{subequations}
\label{eq: constraints}
    \begin{equation}
    \label{eq: dynamic constraints}
        \begin{aligned}
        & \quad \quad  \quad \ \ 
       \tilde{F}_k^{\ast} = \argmin_{\tilde{F}_k}  \|  \tilde{F}_k - \hat{F}_k \|_{\mathcal{F}}^2 \\
        \text{s.t.}  \quad &\text{span}\{ N(\hat{x}_{k+1|k})\} = \text{span}\{  \tilde{F}_k N(\hat{x}_{k|k-1}) \},
        \end{aligned}
    \end{equation}    
and
    \begin{equation}
    \label{eq: measurement constraints}
        \begin{aligned}
           \tilde{H}_k^{\ast} =  &\argmin_{\tilde{H}_k} \|  \tilde{H}_k - \hat{H}_k \|_{\mathcal{F}}^2  \\
            &\text{s.t.} \quad \tilde{H}_k N(\hat{x}_{k|k-1}) = 0.
        \end{aligned}
    \end{equation}    
\end{subequations}
Here $\|\cdot\|_{\mathcal{F}}$ denotes the Frobenius norm. The constraints in \eqref{eq: dynamic constraints} and \eqref{eq: measurement constraints} guarantee that the unobservable subspace is preserved and that the unobservable directions cannot be observed. The observability-constrained EKF algorithm is shown in Alg.~\ref{alg: OC-ESKF}.

\begin{algorithm}[tb!]
\caption{Observability-constrained EKF}\label{alg:alg1}
\begin{algorithmic}
\renewcommand{\algorithmicrequire}{\textbf{Input:}}
\renewcommand{\algorithmicensure}{\textbf{Output:}}
\REQUIRE $\{\bar{u}_k, y_k\}_{k=1}^L$
\ENSURE  $\{\hat{x}_{k|k}, P_{k|k}\}_{k=1}^L$
\\ \textit{Initialisation} : estimated state $\hat{x}_{1|0}$, covariance matrix $P_{1|0}$ 
\STATE {\textbf{For} $k=1$ to $L$ do}
\STATE \hspace{0.5cm}$ \textbf{Measurement update:}$
\STATE \hspace{0.5cm}$ \text{Calculate}\; \hat{H}_k$ using \eqref{eq:F and H normal} and find $\tilde{H}_{k}^{\ast}$ by solving \eqref{eq: measurement constraints}
\STATE \hspace{0.5cm}$ S_{k} = \tilde{H}_{k}^{\ast} P_{k|k-1} (\tilde{H}_{k}^{\ast})^{\top} + R_{k}$
\STATE \hspace{0.5cm}$ K_{k} = P_{k|k-1} (\tilde{H}_{k}^{\ast})^{\top} S_{k}^{-1}$
\STATE \hspace{0.5cm}$ \hat{x}_{k|k} = \hat{x}_{k|k-1} + K_{k} ( y_{k} - h(\hat{x}_{k}))$
\STATE \hspace{0.5cm}$ P_{k|k} = P_{k|k-1} -K_{k} \tilde{H}_{k}^{\ast} P_{k|k-1}$

\STATE \hspace{0.5cm}$ \textbf{Time update:}$
\STATE \hspace{0.5cm}$ \hat{x}_{k+1|k} = f(\hat{x}_{k|k},\bar{u}_k,0) $ 

\STATE \hspace{0.5cm}$ \text{Calculate}\; \hat{F}_k$ using \eqref{eq:F and H normal}  and find $\tilde{F}_{k}^{\ast}$ by solving \eqref{eq: dynamic constraints}
\STATE \hspace{0.5cm}$ G_k = \evalat[\big]{\frac{\partial f}{\partial w_{k}}}{x_{k}=\hat{x}_{k|k}, u_{k}=\bar{u}_{k}, w_k=0}$
\STATE \hspace{0.5cm}$ P_{k+1|k} = \tilde{F}_{k}^{\ast} P_{k|k} (\tilde{F}_{k}^{\ast})^{\top} + G_k Q_k G^{\top}$ 

\STATE {\textbf{end for}}
\end{algorithmic}
\label{alg: OC-ESKF}
\end{algorithm}

\section{Application to the MAINS}
\begin{figure}[t!]
\centering
\includegraphics[trim={0 0 0 30mm},clip,width=0.7\columnwidth]{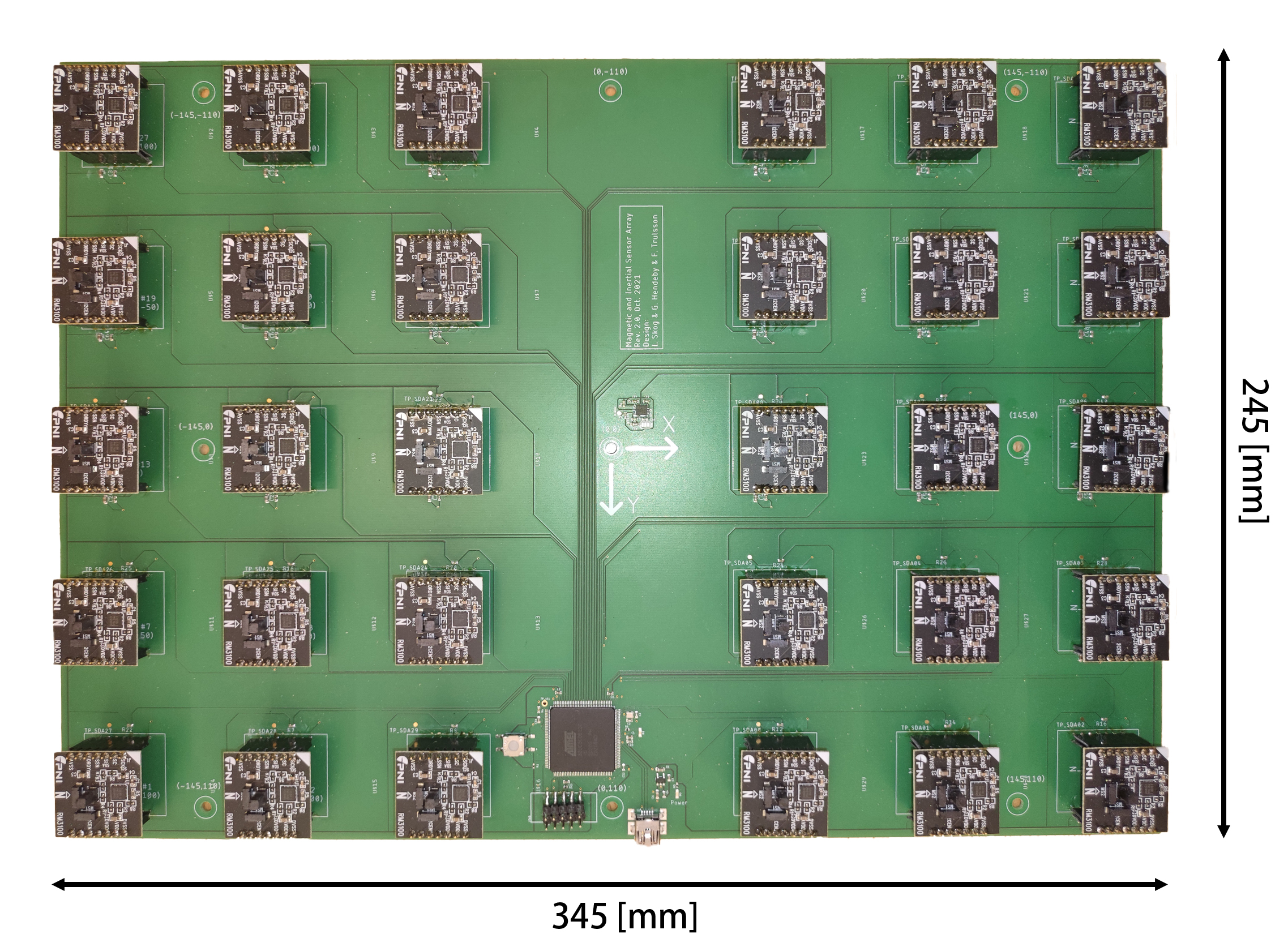}
\caption{The sensor board used in the MAINS. It has 30 PNI \href{https://www.pnicorp.com/rm3100/}{RM3100} magnetometers and an Osmium MIMU 4844 IMU.}
\label{f: sensorboard}
\end{figure}

The MAINS utilizes an IMU and an array of magnetometers, see Fig.~\ref{f: sensorboard}, to perform odometry. The method presented in Section~\ref{chp: preserve estimability} will next be used to modify the EKF algorithm used in the MAINS to address the inconsistency illustrated in Fig.~\ref{f: posterior vs prior uncertainty}. Due to space constraints, the system model presentation next does not include derivations or an in-depth description of the underlying physics. The interested reader is encouraged to read \cite{huang2023mains} for details.     

\subsection{State-Space Model}
Consider a simplified MAINS system in which IMU biases are not included in the state vector. The biases are excluded as they only complicate the forthcoming observability analysis, and biases never improve observability. Later, when evaluating the proposed method, the biases are included.  

Let the state vector $x_k$ be defined as 
\begin{equation}
  x_k \triangleq \begin{bmatrix}
  (p_k^\text{n})^{\top} \;
  (v_k^\text{n})^{\top} \;
  (q_k)^{\top} \;
 (\theta_k)^{\top}
\end{bmatrix}^{\top}. 
\end{equation}
Here, the superscript n denotes the navigation frame where the physical quantity is resolved. Further, $p_k^\text{n} \in \mathbb{R}^3, v^\text{n}_k \in \mathbb{R}^3$ denote the position and velocity, respectively. Moreover, $q_k \in \mathbb{H}$ denotes the unit quaternion that encodes the orientation of the body frame w.r.t. the navigation frame, and $\theta_k \in \mathbb{R}^\kappa$ denotes the coefficients of the polynomial model used to describe the magnetic field; see \cite{huang2023mains} for details.  

The state dynamics is given by \cite{huang2023mains}
\begin{subequations}
  \begin{equation}
    x_{k+1} = f(x_k, u_k, w_k^{\theta}), 
  \end{equation}
where
\begin{align}
  \label{eq: state dynamics}
  f(x_k, u_k, w_k^\theta) &= \begin{bmatrix}
  p_{k}^\text{n} + v_{k}^\text{n} t_\text{s} + (R_k s_k + \text{g}) \frac{t_\text{s}^2}{2} \\
    v_{k}^\text{n} + (R_k s_k + \text{g}) t_\text{s}\\
    q_k \otimes \text{exp}_q\left(\omega_k t_\text{s} \right)\\
    A^{\dagger} B(\psi_k) \theta_k +  w_k^{\theta}
  \end{bmatrix}.
\end{align}
\end{subequations}
Here $u_k \triangleq [s_k^\top \; \omega_k^{\top}]^{\top}$ is the system input, where $s_k \in \mathbb{R}^3$ and $\omega_k \in \mathbb{R}^3$ denote the specific force and angular velocity in body frame, respectively. $w_k^{\theta}\in \mathbb{R}^{\kappa}$ denotes the process noise affecting the coefficients of the magnetic field model. Furthermore, $t_\text{s}$ denotes the sampling period, $R_k\in SO(3)$ denotes the rotation matrix corresponding to $q_k$, $\otimes$ denotes quaternion multiplication, and $\text{exp}_q(\cdot)$ maps an axis-angle rotation to a quaternion. Moreover, the vector $\text{g} \in \mathbb{R}^3$ denotes the local gravity. The matrices $A^{\dagger}$ and $B(\cdot)$ are used in~\cite{huang2023mains} to propagate coefficients of the magnetic field model; see \cite{huang2023mains} for details. Lastly, $\psi_k \in \mathbb{R}^{6}$ denotes the pose change between time $k$ and $k+1$, i.e.,
\begin{equation}
    \psi_k  =\begin{bmatrix}
      R_k^{\top} t_\text{s} (v^n_k + \text{g} t_\text{s}/2)+  s_k t_\text{s}^2/2 \\
      \omega_k t_\text{s}
    \end{bmatrix},
\end{equation}
where the first and second elements encode the translation and orientation change, respectively.

The measurements are from the magnetometer array, whose equation is given by
\begin{subequations}
  \begin{equation}
    y_k = H x_k + e_k,    
  \end{equation}
where
\begin{equation}
  H = \begin{bmatrix}
    0_{3\times10}  & H^{\theta}(r_{1}) \\
    \vdots & \vdots \\
    0_{3\times10}  & H^{\theta}(r_{m})
  \end{bmatrix}.
\end{equation}
\end{subequations}
Here $H^{\theta}(r_{i}) \in \mathbb{R}^{3\times \kappa}$ denotes the measurement matrix for magnetometer sensor location $r_{i}$, whose expression is given in \cite{skog2021magnetic}. Further, $m$ denotes the number of magnetometer sensors.

\subsection{Linearized Error State Model}
Since the state vector contains a quaternion component with a unit norm, the state space is an embedded submanifold \cite{boumal2023introduction}, which is not a vector space. Therefore, the standard derivative typically applied to vector spaces cannot be applied to linearize the model as in \eqref{eq: state dynamics}. Instead, the error state formulation presented in \cite{Joan2017Quaternion} will be used, which uses a small angle approximation to perform the linearization indirectly. 

Let $\delta x_k$ denote the error state vector
\begin{subequations}
    \begin{equation}
    \delta x_k \triangleq  \begin{bmatrix}
  (\delta p_k^\text{n})^{\top} \;
  (\delta v_k^\text{n})^{\top} \;
  (\epsilon_k)^{\top} \;
 (\delta \theta_k)^{\top}
\end{bmatrix}^{\top}.
\end{equation}
Further, let $\bar{x}_k$ denote the nominal state vector
    \begin{equation}
    \bar{x}_k \triangleq  \begin{bmatrix}
  (\bar{p}_k^\text{n})^{\top} \;
  (\bar{v}_k^\text{n})^{\top} \;
  (\bar{q}_k)^{\top} \;
 (\bar{\theta}_k)^{\top}
\end{bmatrix}^{\top}.
\end{equation}
\end{subequations}
Here the standard additive error definition is used, except for the orientation error. For example, $\delta v_k^\text{n} = v^\text{n}_k - \bar{v}_k^\text{n}$. The orientation error $\epsilon_k \in \mathbb{R}^3$ satisfies the equation $R_k\approx \bar{R}_k(I_3 + [\epsilon_k]_{\times})$, where $[\cdot]_{\times}$ is an operator that maps a vector in $\mathbb{R}^3$ to a skew-symmetric matrix such that $\left[\xi\right]_{\times} b = \xi \times b$.

Given the measurement input $\tilde{u}_k = [\tilde{s}_k^{\top}\; \tilde{\omega}_k^{\top}]^{\top}$, the linearized error state dynamics is given by \cite{huang2023mains}
\begin{subequations}
\label{eq: MAINS dynamic equation}
  \begin{equation}
    \delta x_{k+1} = \bar{F}_k \delta x_k + \bar{G}_k w_k,    
  \end{equation}
where
\begin{equation}
      \bar{F}_k \!= \!\begin{bmatrix}
    I_3 \!\!&\!\! I_3 t_\text{s} \!\!&\!\! 0 \!\!&\!\! 0\\
    0   \!\!&\!\! I_3 \!\!&\!\! -\bar{R}_k [\tilde{s}_k]_{\times} t_\text{s} \!\!&\!\! 0 \\
    0   \!\!&\!\! 0  \!\!&\!\! \text{exp}([\tilde{\omega}_k t_\text{s}]_{\times})^{\top} \!\!&\!\! 0 \\
    0   \!\!&\!\! A^{\dagger} \bar{J}_k \bar{R}_k^{\top} t_\text{s} \!\!&\!\! A^{\dagger} \bar{J}_k [\eta(\bar{R}_k, \bar{v}^\text{n}_k)]_{\times} \!\!&\!\! A^{\dagger} B(\bar{\psi}_k) 
  \end{bmatrix},
\end{equation}
\begin{align}
        \bar{\psi}_k &= \begin{bmatrix}
      \bar{R}_k^{\top} t_\text{s} (\bar{v}^\text{n}_k + \text{g} t_\text{s}/2) \\
      \tilde{\omega}_k t_\text{s}
    \end{bmatrix},\\ 
  \eta(\bar{R}_k, \bar{v}^\text{n}_k) &= \bar{R}_k^{\top} t_\text{s}(\bar{v}^{\text{n}}_k+ \text{g} t_\text{s}/2), \\
      \bar{J}_k &= \left. \frac{\partial B(\psi_k) \theta_k}{\Delta p_k}  \right|_{\psi_k=\bar{\psi}_k, \theta_k=\bar{\theta}_k}.
\end{align}
\end{subequations}
Here ${\tilde{s}_k \in \mathbb{R}^3}$ and ${\tilde{\omega}_k \in \mathbb{R}^3}$ denote the measured acceleration and angular velocity, respectively. Further, $w_k\triangleq [(w_k^s)^{\top} \; (w_k^{\omega})^{\top} \;(w_k^{\theta})^{\top}]^{\top}$ denote the process noise, where $w_k^s \in \mathbb{R}^3$ and $w_k^{\omega} \in \mathbb{R}^3$ denote the acceleration and angular velocity measurement noise, respectively. The explicit form of $\bar{G}_k$ is not given as it is relevant in the observability analysis; interested readers can find it in \cite{huang2023mains}. 

The corresponding measurement model is given by 
\begin{subequations}
\label{eq: linearized measurement model}
  \begin{equation}
    \delta y_k = H_{\delta x} \delta x_k + e_k,    
  \end{equation}
where
\begin{equation}
  H_{\delta x} = \begin{bmatrix}
    0_{3\times9}  & H^{\theta}(r_{1}) \\
    \vdots       & \vdots \\
    0_{3\times9}  & H^{\theta}(r_{m})
  \end{bmatrix}
\label{eq: H delta x}
\end{equation}
\end{subequations}
and $\delta y_k \triangleq y_k - H \bar{x}_k$.

\subsection{Unobservable subspace and Interpretations}
Let 
\begin{equation}\label{eq:nullspace base vectors}
    N(x_{k}) \triangleq \begin{bmatrix}
    I_3&0_{3 \times 1}\\
    0_{3\times3}&-[v^{\text{n}}_{k}]_{\times} \text{g} \\
    0_{3\times3}&R_{k}^{\top} \text{g}\\ 
    0_{\kappa \times 3}&0_{\kappa \times 1}         
    \end{bmatrix},
\end{equation}
then the basis of the unobservable subspace associated with the linearized error state model is given by the column vectors in $N(\bar{x}_k)$.
\ifthenelse{\equal{\extendedversion}{1}}{%
See appendix for a proof.
}%
{%
See \cite{HHS2024} for a proof. 
}
The first three columns of $N(\bar{x}_{k})$ correspond to a body frame translation and the last column with the first three corresponds to a navigation frame rotation around the gravity vector. Note that the unobservable velocity component is caused by the fact that the yaw angle cannot be determined. Thus the direction of the velocity in the navigation frame is ambiguous.

\subsection{Suggested Modifications of the Jacobians}
Since the measurement model \eqref{eq: linearized measurement model} is linear and $H_{\delta_x}$ fulfills the constraint in \eqref{eq: measurement constraints}, it is left unmodified, i.e., $\tilde{H}_k^{\ast} = H_{\delta_x}$. Only the Jacobian involved in the state transition, i.e.,
\begin{equation}
     \hat{F}_k\!=\!\begin{bmatrix}
    I_3\!\!\!&\!\!\!I_3 t_\text{s} \!\!&\!\! 0 \!\!&\!\! 0\\
    0&I_3 \!\!&\!\! -\hat{R}_k [\tilde{s}_k]_{\times} t_\text{s} \!\!&\!\! 0 \\
    0\!\!&\!\!0  \!\!&\!\!\text{exp}([\tilde{\omega}_k t_\text{s}]_{\times})^{\top} \!\!&\!\! 0 \\
    0\!\!&\!\!A^{\dagger}\!\hat{J}_k \hat{R}_k^{\top} t_\text{s}\!\!&\!\! A^{\dagger}\!\hat{J}_k \hat{R}_k^{\top} t_\text{s} [\hat{v}^{\text{n}}_k+\text{g}\frac{t_\text{s}}{2}]_{\times} \hat{R}_k\!\!&\!\!\hat{T}_k^{k+1}
    \end{bmatrix}
\end{equation}
needs to be modified. Here the accent $\hat{\cdot}$ denotes the posterior estimate of the quantity and $\hat{T}_k^{k+1} \triangleq A^{\dagger} B(\hat{\psi}_k)$. Furthermore, the equallity $[R \xi]_{\times} = R [\xi]_{\times} R^{\top}, R\in SO(3), \xi \in \mathbb{R}^3$ is used.

Solving the optimization problem in \eqref{eq: dynamic constraints} is typically difficult, However, a suboptimal solution can be obtained by narrowing the search space for $\tilde{F}_k$ and transforming the constraint by selecting a specific set of transformations $\{\mathcal{E}_k, \mathcal{E}_{k+1}\}$ such that the basis vectors in the constraint are equal, i.e.,
\begin{equation}
\label{eq: equivalent constraint}
N(\hat{x}_{k+1|k}) \mathcal{E}_{k+1} =\tilde{F}_k N(\hat{x}_{k|k-1}) \mathcal{E}_k.    
\end{equation}

Since the goal is to make minimal changes to the original Jacobian $\hat{F}_k$ but still fulfill the constraints in \eqref{eq: dynamic constraints}, the sub-blocks of matrices $\hat{F}_k$ that are independent of the linearization points are kept unchanged. Furthermore, the sub-block $\hat{T}_k^{k+1}$ will not affect the constraint since the last $\kappa$ rows of $N(\hat{x}_{k|k-1})$ are all zeros. Hence, it is kept as it is. The remaining subblocks of $\hat{F}_k$ must be modified to meet the constraint in \eqref{eq: dynamic constraints}. Therefore, the proposed modified Jacobian $\tilde{F}_k$ has the structure 
\begin{equation}
  \tilde{F}_k=\begin{bmatrix}
    I_3\!&\!I_3 t_\text{s} & 0 & 0\\
    0\!&\!I_3 \!&\! \tilde{F}_k^{(1,2)} \!&\! 0 \\
    0\!&\!0  \! &\! \tilde{F}_k^{(2,2)} \!&\! 0 \\
    0\!&\!A^{\dagger} \hat{J}_k \hat{R}_k^{\top} t_\text{s}\!&\! A^{\dagger} \hat{J}_k \hat{R}_k^{\top} t_\text{s} \tilde{F}_k^{(3,2)}\! & \!\hat{T}_k^{k+1}
  \end{bmatrix},
\end{equation}
where $\tilde{F}_k^{(1,2)}$, $\tilde{F}_k^{(2,2)}$, and $\tilde{F}_k^{(3,2)}$ are the block matrices to be determined. Further, transformations $\{\mathcal{E}_k, \mathcal{E}_{k+1}\}$ are chosen as
\begin{equation}
    \mathcal{E}_k = \begin{bmatrix}
        I_3 & a_k \\
        0   & 1
    \end{bmatrix} , \quad
    \mathcal{E}_{k+1} = \begin{bmatrix}
        I_3 & a_{k+1} \\
        0   & 1
    \end{bmatrix} . \quad
\end{equation}
Here $a_k, a_{k+1} \in \mathbb{R}^3$ are column vectors.
Then the optimization problem \eqref{eq: dynamic constraints} can be written as
\begin{subequations}\label{eq:mains opt}
\begin{align}
 \tilde{F}_k^{\ast}&=\argmin_{\tilde{F}_k}  \|  \tilde{F}_k - \hat{F}_k \|_{\mathcal{F}}^2\\
\label{eq: constraints T}
     \text{s.t.}\quad &a_{k+1}= a_k -[\hat{v}^{\text{n}}_{k|k-1}]_{\times} \text{g} t_{\text{s}},\\ 
     \label{eq: constraint 1}
     &[\hat{v}^{\text{n}}_{k+1|k}]_{\times} \text{g} =[\hat{v}^{\text{n}}_{k|k-1}]_{\times} \text{g} - \tilde{F}_k^{(1,2)} \hat{R}_{k|k-1}^{\top} \text{g}, \\
        &\hat{R}_{k+1|k}^{\top} \text{g}  =  \tilde{F}_k^{(2,2)} \hat{R}_{k|k-1}^{\top}  \text{g}\\
        \label{eq: constraints b}
     &A^{\dagger} \hat{J}_k \hat{R}_k^{\top} t_\text{s} (-[\hat{v}^{\text{n}}_{k|k-1}]_{\times}\text{g} +  \tilde{F}_k^{(3,2)} \hat{R}_{k|k-1}^{\top}\text{g})=0   
\end{align}
\end{subequations}
These constraints are obtained by matching the entries on both sides of \eqref{eq: equivalent constraint}.

The constraint \eqref{eq: constraints T} does not concern the modification of the Jacobian. It is the result of keeping the first row of $\tilde{F}_k$ the same as that of $\hat{F}_k$. 

Since the constraint in \eqref{eq:mains opt} only contains one unique subblock of $\tilde{F}_k$ the optimization problem in \eqref{eq:mains opt} can be split into three separate optimization problems of the form
\begin{equation}
\begin{split}
    \tilde{F}^{\ast^{(i,2)}} = &\argmin_{\tilde{F}^{(i,2)}}\|\tilde{F}^{(i,2)} - \hat{F}^{(i,2) }\|_{\mathcal{F}}^2,\\
    \text{s.t.} \quad &  \tilde{F}^{(i,2)} u = w.
\end{split}
\end{equation}
Here $\hat{F}^{(i,2)}$ denotes the matrix to be modified and $u, w \in \mathbb{R}^3$. For example, the constraint \eqref{eq: constraint 1} corresponds to $u= \hat{R}_{k|k-1}^{\top} \text{g}$, $w=[\hat{v}^{\text{n}}_{k|k-1} - \hat{v}^{\text{n}}_{k+1|k}]_{\times} \text{g}$. The optimization problem has a closed-form solution \cite{Hesch2014Consistency}
\begin{equation}
    \tilde{F}^{\ast^{(i,2)}} =  \hat{F}^{(i,2)} - (\hat{F}^{(i,2)} u - w) (u^{\top}u)^{-1} u^{\top}.
\end{equation}
Note that in \cite{Hesch2014Consistency}, the optimization problem set for $\tilde{F}^{(2,2)}$ is different, where $\tilde{F}^{(2,2)}$ is constrained to be a rotation matrix and the object is to minimize the squared norm of the difference of the quaternions corresponding to the rotation matrices. In this paper, we also adopted this approach. Furthermore, when dealing with \eqref{eq: constraints b}, we consider the term in the parenthesis to be 0, although in the general case, it can be any vector in the nullspace of $A^{\dagger} \hat{J}_k \hat{R}_k^{\top} t_\text{s}$.

\section{Experimental Evaluation}
The proposed observability-constrained MAINS (OC-MAINS) algorithm is compared with the original MAINS algorithm in \cite{huang2023mains}. Both algorithms are evaluated using Monte Carlo simulations and real-world data. As a performance measure, the root mean squared error (RMSE), defined as
\begin{equation}
    \text{RMSE} = \left(\frac{1}{M} \sum_{i=1}^M \| \hat{\rho}_{k,i}-\rho_{k,i}\|^2 \right)^{1/2},
\end{equation}
is used. Here, $\rho$ denotes the quantity for which RMSE is computed, $M$ is the number of simulations, and $i$ denotes the simulation index. Furthermore, the perceived uncertainty of the estimate at each timestep is calculated as
\begin{equation}
    \text{perceived uncertainty}= \left(\frac{1}{M} \sum_{i=1}^M P^{\rho}_{k, i}\right)^{1/2}.
\end{equation}
Here $P^{\rho}_{k, i}$ denotes the covariance of the posterior estimate of the quantity $\rho$ calculated by the EKF during the $i$:th simulation. The covariance of the yaw is calculated as 
\begin{equation}
       P^{\phi}_{k, i} = {\nabla_{\text{yaw}}(\hat{q}_{k,i})}^{\top} P^{\epsilon}_{k, i} \nabla_{\text{yaw}}(\hat{q}_{k,i}).
\end{equation}
Here $P^{\epsilon}_{k, i} \in \mathbb{R}^{3\times3}$ denote the block covariance matrix corresponding to the orientation error. Furthermore, $\nabla_{\text{yaw}}(\hat{q}_{k,i}) \in \mathbb{R}^{3 \times 1}$ denotes the gradient w.r.t the Euler angles of the function that converts $\hat{q}_{k,i}$ to the yaw angle.

\subsection{Simulation Setup and Result} \label{section: simulation}
In the simulation, a sensor board, similar to that shown in Fig.~\ref{f: sensorboard}, moves in squares of size 2 $\times$ 2 meters; see Fig.~\ref{fig: simulated trajectory}. The trajectory's duration was 8 seconds, and the data was sampled at 100 Hz. The IMU measurements were generated as the true value corrupted by white noise and biases. The magnetometer measurements were generated as the magnetic field from a multi-dipole model corrupted by additive white noise. The magnitude of the simulated magnetic field is in Fig.~\ref{fig: simulated trajectory}. In total, 50 simulations with independent noise, bias, and initial state realizations were used in the Monte Carlo simulation evaluation. The same parameter settings were used in both algorithms. Due to space constraints, all the settings are not presented here, but they can be found in the published code.      

The results are shown in Fig.~\ref{f: position RMSE simulation} and Fig.~\ref{f: yaw error simulation}. The figures show that the proposed OC-MAINS algorithm generally has a smaller position and yaw RMSE. Further, the perceived uncertainty of the yaw estimate is more consistent with the true uncertainty (See also Fig.~\ref{f: posterior vs prior uncertainty}.). Concerning the position estimates, even though the perceived uncertainty of the OC-MAINS is somewhat more consistent with the true uncertainty of the position estimates, significant inconsistency still exists. However, it is worth noting that the perceived uncertainty of OC-MAINS is always higher than the initial uncertainty, thus satisfying the condition in \eqref{eq:inequalities}. The original MAINS algorithm does not meet this condition.  

\subsection{Experimental Setup and Result}
In the experiment, a person held the sensor board in Fig.~\ref{f: sensorboard} parallel to the ground and walked in squares for a few laps. The true trajectory was measured using a camera-based tracking system. The same parameters were used in both algorithms.  

\begin{figure*}[t!]
        \centering
        \begin{minipage}{0.47\textwidth}
        \begin{subfigure}[b]{\columnwidth}
            \centering
            \includegraphics[width=\linewidth]{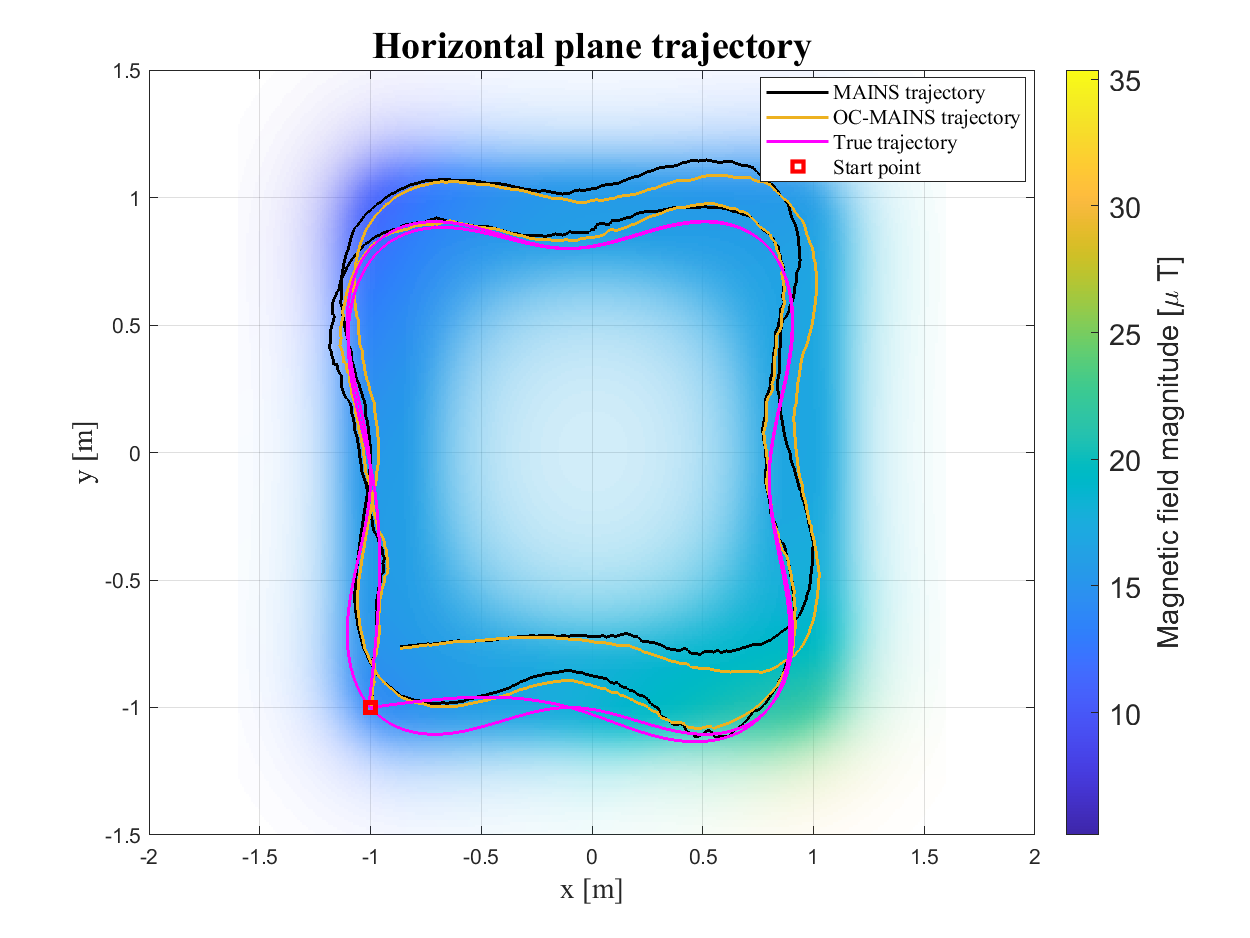}
            \caption{Estimated and true trajectory, and the magnetic field magnitude.\newline}
            \label{fig: simulated trajectory}
        \end{subfigure}
        \hfill
        \begin{subfigure}[b]{\columnwidth}
            \centering
            \includegraphics[width=\linewidth]{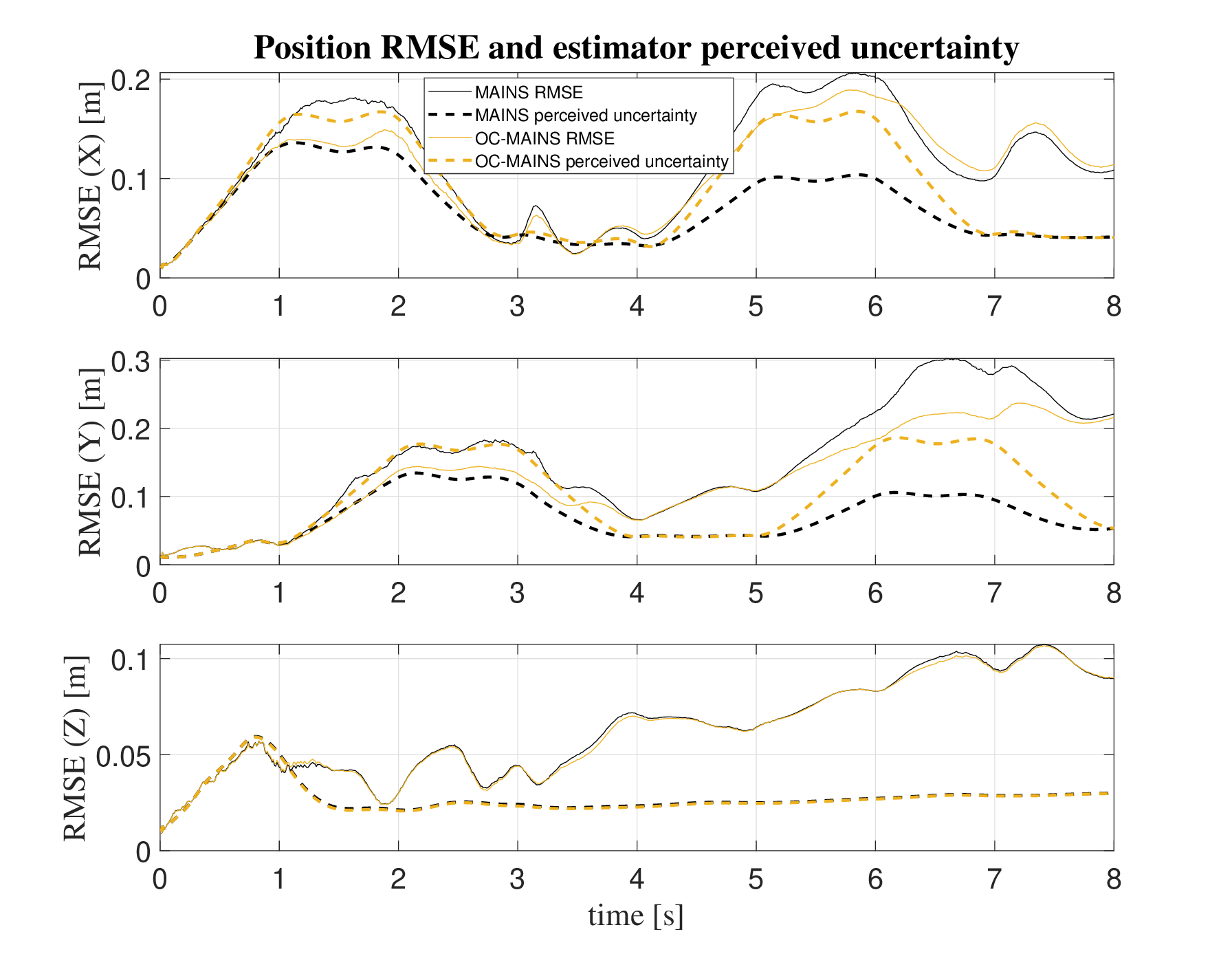}
            \caption{Position RMSE (solid) and perceived uncertainty (dashed).}
            \label{f: position RMSE simulation}
        \end{subfigure}
        \vskip\baselineskip
        \begin{subfigure}[b]{\columnwidth}
            \centering
            \includegraphics[width=\linewidth]{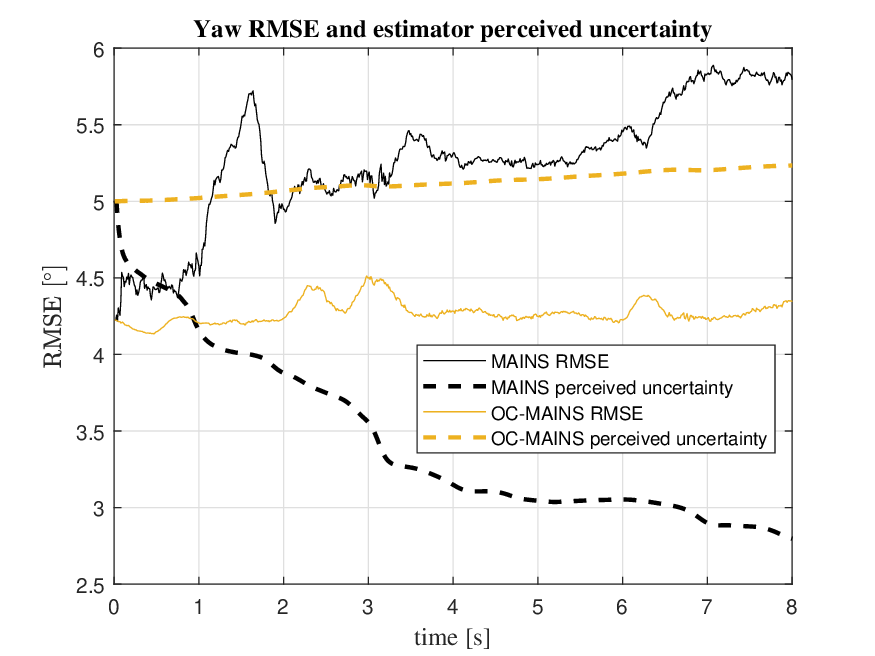}
            \caption{Yaw RMSE (solid) and perceived uncertainty (dashed).}
            \label{f: yaw error simulation}
        \end{subfigure}
        \caption{Monte Carlo simulation results. The RMSE results are the average value calculated from 50 independent simulations.}
        \label{fig:4subfigures_1}
  \end{minipage}
  \hfill
  \begin{minipage}{0.47\textwidth}
        \begin{subfigure}[b]{\columnwidth}
            \centering
            \includegraphics[width=\linewidth]{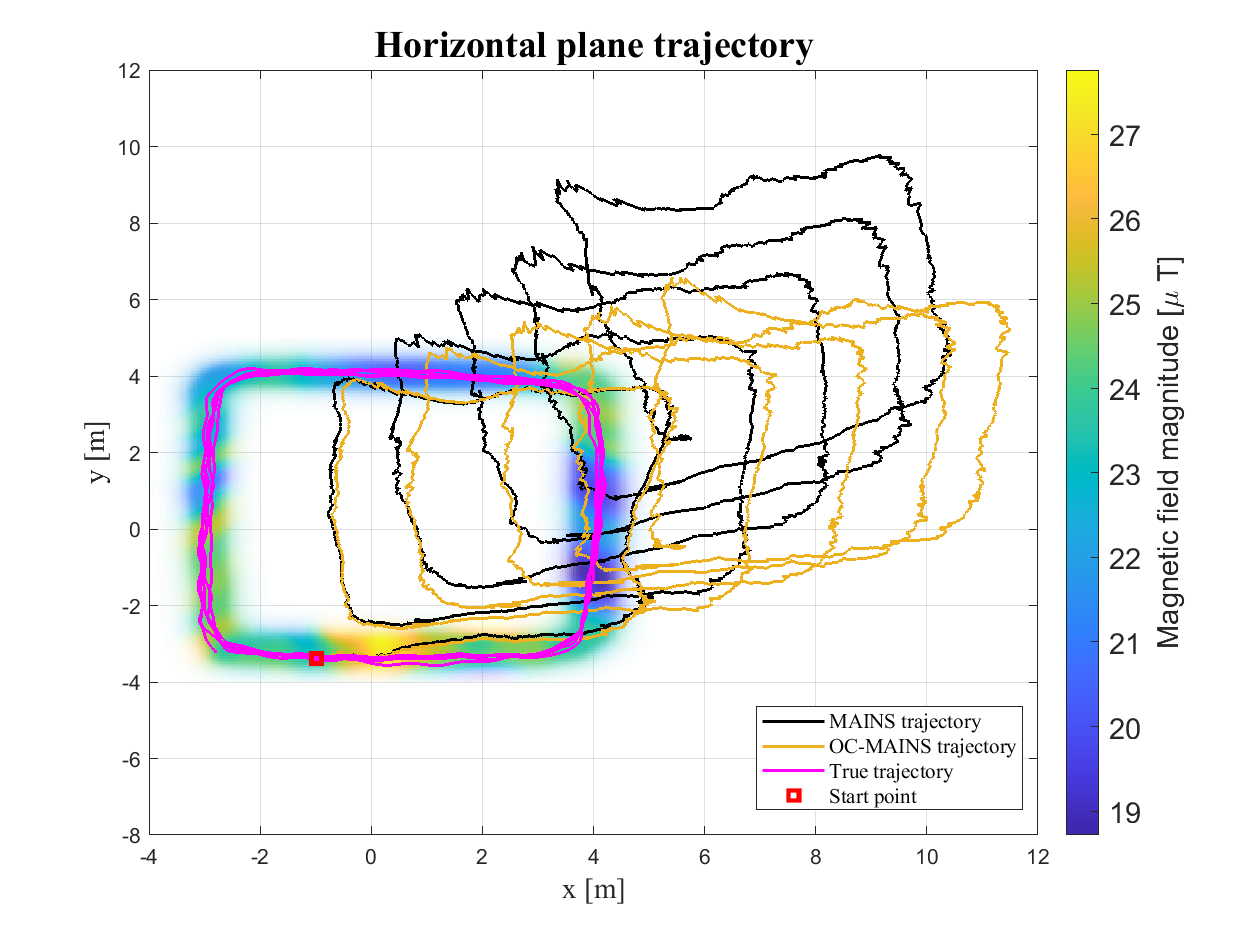}
            \caption{Estimated and true trajectory, and the magnetic field magnitude}
            \label{f: Real trajectory}
        \end{subfigure}
        \hfill
        \begin{subfigure}[b]{\columnwidth}
            \centering
            \includegraphics[width=\linewidth]{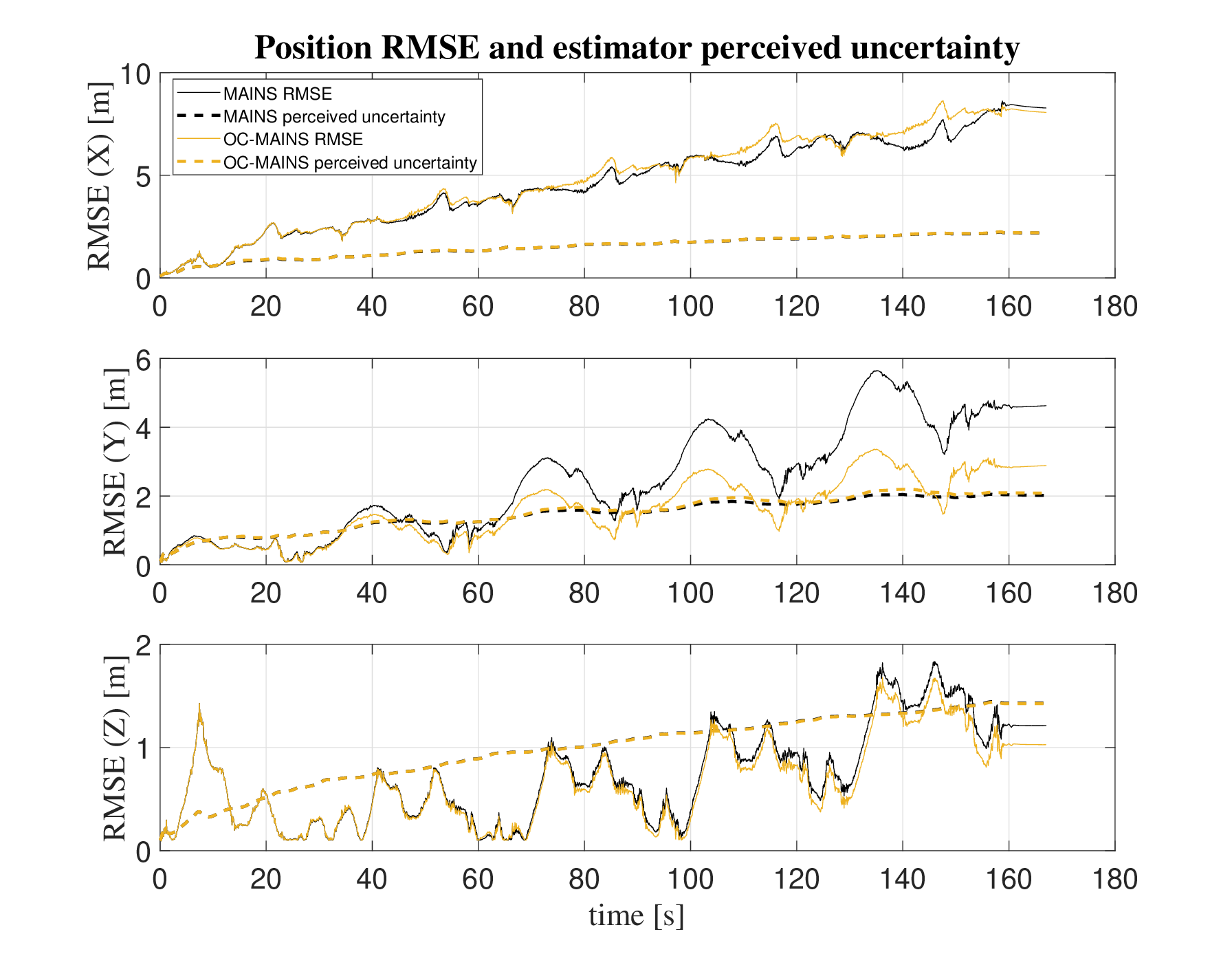}
            \caption{Position RMSE (solid) and perceived uncertainty (dashed).}
            \label{f: position RMSE Real}
        \end{subfigure}
        \vskip\baselineskip
        \begin{subfigure}[b]{\columnwidth}
            \centering
            \includegraphics[width=\linewidth]{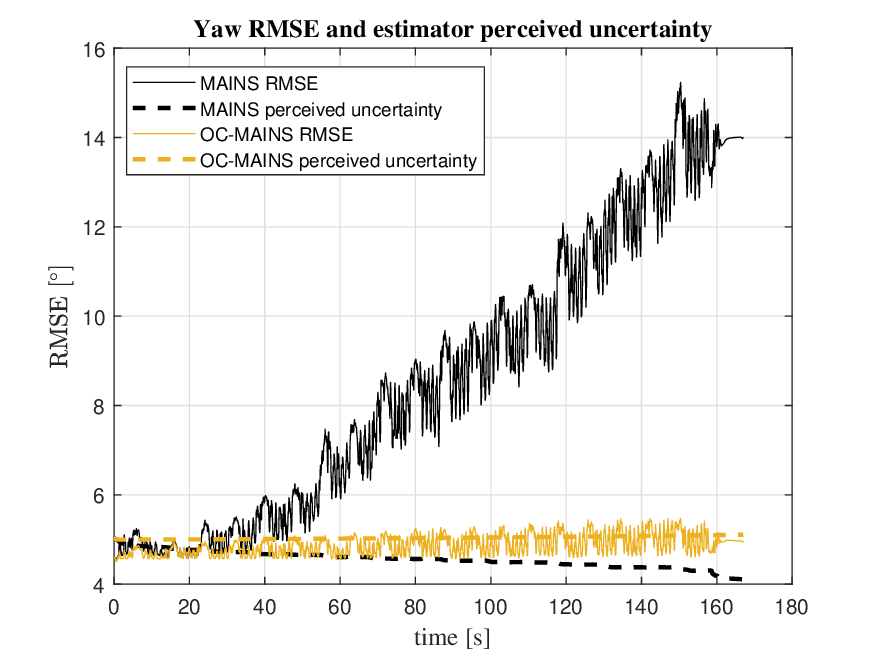}
            \caption{Yaw RMSE (solid) and perceived uncertainty (dashed).}
            \label{f: yaw error Real}
        \end{subfigure}
        \caption{Results from real-world experiment. The RMSE results are the average value calculated from using 12 randomly selected initialization states.}
  \end{minipage}
    \end{figure*}

The results are shown in Fig.~\ref{f: position RMSE Real} and Fig.~\ref{f: yaw error Real}. From Fig.~\ref{f: yaw error Real}, it can be seen that the yaw error of OC-MAINS is significantly lower than that of the original MAINS. Also, it can be seen that the perceived and true yaw uncertainty agree when using the OC-MAINS algorithm. From Fig.~\ref{f: position RMSE Real}, it can be seen that the position error of the original MAINS and OC-MAINS algorithms are about the same, but the OC-MAINS algorithm has slightly better performance in the $y$-axis direction. For the OC-MAINS algorithm, the perceived and true uncertainty agrees in the $y$- and $z$-axis directions, whereas in the $x$-axis, there is a significant inconsistency. This could result from imperfect IMU calibration. Similar to in the simulations, the perceived uncertainty of OC-MAINS is always higher than the initial uncertainty, thus satisfying the condition in \eqref{eq:inequalities}. 

\section{Summary \& Conclusions}
A method to construct an observability-constrained magnetic field-aided INS has been proposed. The proposed method modifies the state transition matrix in the EKF to ensure that the observability properties of the nonlinear system are preserved in the filtering process. The method extends the work in  \cite{Hesch2014Consistency} by introducing a transformation of the basis that spans the unobservable subspace. This transformation reduces the changes that must be done to the state transition matrix. The proposed method was evaluated on simulation and real-world datasets, showing that (i) the observability properties are preserved, (ii) the estimation accuracy increases, and (iii) the perceived uncertainty calculated by the EKF is more consistent with the true uncertainty of the estimates.   

\section*{Acknowledgment}
This work has been funded by Security Link and the Swedish Research Council project 2020-04253.

\IEEEpeerreviewmaketitle

\ifthenelse{\equal{\extendedversion}{1}}{
\input{appendixfile}
}
{}

\bibliographystyle{IEEEtran}
\bibliography{IEEEabrv,ref_ipin2024}

\end{document}

%% file: appendixfile.tex
\appendices
\section{The Unobservable Subspace Basis}
To derive the unobservable subspace basis for the system defined by \eqref{eq: MAINS dynamic equation} and \eqref{eq: linearized measurement model}, $\bar{F}_k$ is rewritten. Propagating the nominal state $\bar{x}_k$ with $\tilde{u}_k$ using \eqref{eq: state dynamics} leads to
\begin{subequations}
\begin{align}
    \bar{v}^{\text{n}}_{k+1}
    &= \bar{v}^{\text{n}}_{k} + \bar{R}_k \tilde{s}_k t_\text{s} + \text{g} t_\text{s}, \\
    \bar{q}_{k+1}
    &= \bar{q}_k \otimes \text{exp}_q\left( \tilde{\omega}_k t_\text{s} \right).
\end{align}
\end{subequations}
By rearranging the terms and using the equality $[R \xi]_{\times} = R [\xi]_{\times} R^{\top}$, where $R \in SO(3)$ and $\xi \in \mathbb{R}^3$, the following holds
\begin{subequations}\label{eq: F replacement 1}
  \begin{align}
    \bar{R}_k [\tilde{s}_k]_{\times} t_\text{s}
    &= [\bar{v}^{\text{n}}_{k+1} - \bar{v}^{\text{n}}_k - \text{g} t_\text{s}]_{\times} \bar{R}_k, \\
    \text{exp}([\tilde{\omega}_k t_\text{s}]_{\times})^{\top}
    &= \bar{R}_{k+1}^{\top} \bar{R}_k ,
  \end{align}
\end{subequations}
Let $\bar{T}_{k}^{k+1} \triangleq A^{\dagger} B(\bar{\psi}_k)$ and $\Delta \bar{p}_k^{G} =\bar{R}_k \Delta \bar{p}_k$, then it holds that
\begin{subequations}
\label{eq: F replacement 2}
\begin{equation}
A^{\dagger} \bar{J}_k \bar{R}_k^{\top} t_\text{s} = \frac{\partial A^{\dagger} B(\bar{\psi}_k) \bar{\theta}_k}{\partial \Delta \bar{p}_k} \bar{R}_k^{\top} t_\text{s}=
\frac{\partial \bar{T}_{k}^{k+1} \bar{\theta}_k}{\partial \Delta \bar{p}_k^{G}} t_\text{s} ,
\end{equation}
and
\begin{equation}\label{eq: del theta del epsilon}
  A^{\dagger} \bar{J}_k [\eta(\bar{R}_k, \bar{v}^{\text{n}}_k)]_{\times}  =
 \frac{\partial \bar{T}_{k}^{k+1} \bar{\theta}_k}{\partial \Delta \bar{p}_k^{G}} t_\text{s} [\bar{v}^{\text{n}}_k+\text{g} t_\text{s}/2]_{\times}\bar{R}_k.
\end{equation}
\end{subequations}
Next, combining \eqref{eq: F replacement 1} and \eqref{eq: F replacement 2}, then $\bar{F}_k$ can be rewritten as 
    \begin{IEEEeqnarray}{rl}
&\bar{F}_k\nonumber \\
&= \begin{bmatrix}
    I_3 \!&\! I_3 t_\text{s} \!&\! 0 \!&\! 0\\
    0 \!&\! I_3 \!&\! -[\bar{v}^{\text{n}}_{k+1} - \bar{v}^{\text{n}}_k - \text{g} t_\text{s}]_{\times} \bar{R}_k \!&\! 0 \\
    0 \!&\! 0  \!&\! \bar{R}_{k+1}^{\top} \bar{R}_k \!&\! 0 \\
    0 \!&\!\frac{\partial \bar{T}_{k}^{k+1} \bar{\theta}_k}{\partial \Delta \bar{p}_k^{G}} t_\text{s} \!&\! \frac{\partial \bar{T}_{k}^{k+1} \bar{\theta}_k}{\partial \Delta \bar{p}_k^{G}} t_\text{s} [\bar{v}^{\text{n}}_k+\text{g} t_\text{s}/2]_{\times}\bar{R}_k \!&\!  \bar{T}_{k}^{k+1}
  \end{bmatrix}, \nonumber \\*
\end{IEEEeqnarray}
The matrix $\bar{\Phi}(l,k)$ can then, for $l > k$, be written as\begin{subequations}
 \begin{IEEEeqnarray}{rl}
    &\bar{\Phi}(l,k)   \nonumber \\
     &= \begin{bmatrix}
      I_3 \!&\! \bullet \!& \! \bullet \!&\! 0 \\
      0 \!&\! I_3 \!&\! -[\bar{v}^{\text{n}}_{l}-\bar{v}^{\text{n}}_{k}-\text{g}(l-k)t_\text{s}]_{\times} \bar{R}_{k} \!&\! 0 \\
      0 \!&\! 0 \!&\! \bar{R}_{l}^{\top} \bar{R}_{k} \!&\! 0 \\
      0 \!&\! \sum\limits_{j=k}^{l-1} D_j  \!&\! \sum\limits_{j=k}^{l-1} D_j [\bar{v}^{\text{n}}_{k}+\frac{2(j-k)+1}{2} t_\text{s} \text{g} ]_{\times} \bar{R}_{k}\!&\! \bar{T}^l_{k}
    \end{bmatrix}, \nonumber \\*
    \end{IEEEeqnarray}
where
\begin{equation}
      D_j = \frac{\partial \bar{T}^l_{k} \bar{\theta}_{k}}{\partial \Delta \bar{p}_j^{G}} t_\text{s}\quad\text{and}\quad 
    \bar{T}^l_{k} = \bar{T}^l_{l-1} \bar{T}^{l-1}_{l-2} \cdots \bar{T}^{k+1}_{k}.
\end{equation}
\end{subequations}
Here $\bullet$ denotes subblocks not required for the analysis.

Now, it can be verified that the columns of $N(\bar{x}_k)$ in \eqref{eq:nullspace base vectors} are unobservable error states by multiplying every block row in $\bar{\mathcal{O}}_{k}$ with $N(\bar{x}_k)$. Let $\bar{\mathcal{O}}_{k}^{(s,l)} \triangleq H_{\delta x}^{(s)} \bar{\Phi}(l,k)$, where $H_{\delta x}^{(s)}$ denotes the block row in $H_{\delta x}$ corresponding to the $s$-th magnetometer. For $l=k$, \begin{IEEEeqnarray}{rCl}
    \bar{\mathcal{O}}_{k}^{(k,k)}=\begin{bmatrix}
        0_{3\times3} & 0_{3\times3} & 0_{3\times3} & H^{\theta}(r_s)
    \end{bmatrix},
\end{IEEEeqnarray}
and $\bar{\mathcal{O}}_{k}^{(s,k)} N(\bar{x}_{k}) = \left[0_{3\times3} \quad 0_{3\times 1} \right]$, which can be verified by simple calculation using that $\bar{\Phi}(k,k)=I$.

For $l>k$,
\begin{equation}
        \bar{\mathcal{O}}_{k}^{(s,l)}  = H^{\theta}(r_{s})\!\begin{bmatrix}
          0_{3\times3}\\
      (\sum \limits_{j=k}^{l-1} D_j)^{\top} \\
      (\sum \limits_{j=k}^{l-1} D_j [\bar{v}^{\text{n}}_{k}+\frac{2(j-k)+1}{2} t_\text{s} \text{g}]_{\times} \bar{R}_{k})^{\top}\\
      (\bar{T}^l_{k})^{\top}
    \end{bmatrix}^{\top}
  \end{equation}
and 
  \begin{IEEEeqnarray}{rCl}
          \bar{\mathcal{O}}_{k}^{(s,l)} N(\bar{x}_{k})  &=&   H^{\theta}(r_{m_s})  \left[ 0 \quad \sum \limits_{j=k}^{l-1} D_j (-[\bar{v}^{\text{n}}_{k}]_{\times}\text{g} \right. \nonumber\\ 
          &&+ \> \left.[\bar{v}^{\text{n}}_{k} + \frac{2(j-k)+1}{2} t_\text{s} \text{g} ]_{\times} \bar{R}_{k} \bar{R}_{k}^{\top} \text{g}) \right] \nonumber\\ 
          &=& H^{\theta}(r_{m_s}) \left[0 \quad \sum \limits_{j=k}^{l-1} D_j (-[\bar{v}^{\text{n}}_{k}]_{\times}\text{g} + [\bar{v}^{\text{n}}_{k}]_{\times} \text{g} \right.\nonumber\\
          &&+ \left. \> [\frac{2(j-k)+1}{2} t_\text{s} \text{g} ]_{\times}\text{g}) \right]\nonumber\\
          &&=\left[0_{3\times3} \quad 0_{3\times 1} \right], 
  \end{IEEEeqnarray}
Here the equality $[\text{g}]_{\times} \text{g} = 0$ is used in the last step.

Since $\mathcal{\bar{O}}_{k}^{(s,l)} N(\bar{x}_{k}) = 0$ $\forall s,l\geq k$ then $\mathcal{\bar{O}}_{k} N(\bar{x}_{k}) = 0$ and the column vectors of $N(\bar{x}_{k})$ belong to the null space of $\bar{\mathcal{O}}_{k}$.

\section{Interpretation of the Unobservable Subspace}
The error state vector can be seen as perturbations to the state vector. The proof proceeds by applying a perturbation to the state and then calculating the resulting error state vector. If the error state vector lies in the space spanned by some columns in $N(\bar{x}_{k})$ then those unobservable directions correspond to the perturbation applied.

Suppose the initial true state $x=[p^{\text{n}^\top}\; v^{\text{n}^\top}\; q^{\top}\; \theta^{\top}]^{\top}$, where the time index will be dropped for brevity. Next, suppose that a translation $\Delta \in \mathbb{R}^3$ is applied to the body frame. That will change the state to $x^{\prime}=[(p^{\text{n}}\!+\!\Delta)^\top \;v^{\text{n}^\top}\; q^{\top}\; \theta^{\top}]^{\top}$. The error state $\delta x = [-\Delta^{\top} \quad 0^{\top} \quad 0^{\top} \quad 0^{\top}]^{\top}$ lies in the space spanned by the first three columns in $N(\bar{x}_{k})$. Thus, the perturbation implied by the first three columns is a body frame translation.

Next, suppose the navigation frame is rotated around the gravity vector by an angle $c \in [0,2\pi)$ and denote the new navigation frame as $\text{n}^{\prime}$-frame, then the changed state $x^{\prime}$ becomes
\begin{equation}
        x^{\prime}=\begin{bmatrix}
       {}_\text{n}^{\text{n}^{\prime}}R \cdot p^\text{n}\\
       {}_\text{n}^{\text{n}^{\prime}}R \cdot v^\text{n}\\
        q\{{}_\text{n}^{\text{n}^{\prime}}R\} \otimes q \\
        \theta\\
    \end{bmatrix}, \quad
        {}_\text{n}^{\text{n}^{\prime}}R = R^{\top}\{c\,\text{g}\},
\end{equation}
where  ${}_\text{n}^{\text{n}^{\prime}}R$ denotes the rotation matrix that transforms the coordinate in $\text{n}$-frame to $\text{n}^{\prime}$-frame, $R\{c\,\text{g}\}$ denotes the rotation matrix corresponding to the axis-angle representation $c\,\text{g}$, and  $q\{ {}_\text{n}^{\text{n}^{\prime}}R\}$ denotes the quaternion corresponding to the rotation matrix ${}_\text{n}^{\text{n}^{\prime}}R$.

Let $\epsilon \in \mathbb{R}^3$ denote the orientation error, i.e. the orientation difference in $x$ and $x^{\prime}$, and $c$ be infinitesimally small. Recalling the definition of orientation error and using $R^{\top}\{c\,\text{g}\} \approx I_3 -  [c\,\text{g}]_{\times}$, it holds that
\begin{equation}
\begin{split}
    R &\approx   ({}_\text{n}^{\text{n}^{\prime}}R \cdot R ) \cdot  (I_3 + [\epsilon]_{\times}) = (R^{\top}\{c\,\text{g}\} \cdot R ) \cdot  (I_3 + [\epsilon]_{\times}) \\
    &\approx  (I_3 -  [c\,\text{g}]_{\times})  R  (I_3 + [\epsilon]_{\times})
\end{split}
\end{equation}
where $R$ is the rotation matrix corresponding to the quaternion $q$. Neglecting the second-order term, then it holds that    
\begin{equation}
    R \approx   R + R  [\epsilon]_{\times} -  [c\,\text{g}]_{\times}  R,
\end{equation}
which leads to 
\begin{equation}
    [c \, \text{g}]_{\times} \approx R [\epsilon]_{\times} R^{\top} = [R \epsilon]_{\times}.
\end{equation}
Therefore,
\begin{equation}
 \epsilon = c R^{\top}\text{g}.   
\end{equation}
The error state for the position is calculated as 
\begin{equation}
\begin{split}
    p^{\text{n}} - {}_\text{n}^{\text{n}^{\prime}}R \cdot p^\text{n} &= p^{\text{n}} - R^{\top}\{c\,\text{g}\}\cdot p^\text{n} \\
   & \approx (I_3 - (I_3 - [c \, \text{g}]_{\times})) \cdot  p^{\text{n}} ,\\
   & = -c [p^{\text{n}} ]_{\times} \text{g}
\end{split}
\end{equation}
where $[\text{g}]_{\times} p^\text{n} = -[p^\text{n}]_{\times}\text{g}$ is used. Similarly, the error state for the velocity is $-c [v^{\text{n}} ]_{\times} \text{g}$. Therefore, the error state is
\begin{equation}
 \delta x = c\begin{bmatrix}
     -[p^\text{n}]_{\times}\text{g}\\
     -[v^\text{n}]_{\times}\text{g}\\
     R^{\top}\text{g}\\
     0
 \end{bmatrix} = c \left(-\begin{bmatrix}
     I_3\\
    0_{3\times3}\\
    0_{3\times3}\\ 
    0_{\kappa \times 3}   
 \end{bmatrix}  [p^\text{n}]_{\times}\text{g} + \begin{bmatrix}
     0\\
     -[v^\text{n}]_{\times}\text{g}\\
     R^{\top}\text{g}\\
     0
 \end{bmatrix}\right)
\end{equation}
lies in the space spanned by all columns, so they correspond to a navigation frame rotation around the gravity vector.